\documentclass{article}

\PassOptionsToPackage{numbers, compress}{natbib}
\usepackage[main, preprint]{neurips_2026}

\usepackage{graphicx}
\usepackage{booktabs}
\usepackage{hyperref}
\usepackage{multirow,makecell,wrapfig,algorithm,algpseudocode}
\usepackage{bbding,amsmath,bbm,graphicx,subcaption,amsfonts,pifont,mathtools,booktabs,diagbox,tikz,xcolor,bm,colortbl,enumitem}

\definecolor{artifact}{HTML}{FF887B}
\definecolor{zoom}{HTML}{B4E3FF}
\definecolor{Gold}{RGB}{255,215,0}
\definecolor{Silver}{RGB}{192,192,192}
\definecolor{Bronze}{RGB}{205,127,50}
\definecolor{red}{RGB}{255,0,0}
\colorlet{GoldSoft}{Gold!25}
\colorlet{SilverSoft}{Silver!25}
\colorlet{BronzeSoft}{Bronze!25}

\usepackage{mdframed}
\usepackage{xcolor}

\definecolor{red}{RGB}{255,0,0}

\definecolor{green}{RGB}{0,255,0}

\title{A Causal Diffusion Model for Video Reconstruction from Ultra-Low-Bitrate Representations}

\author{%
  \textbf{Cem Eteke}$^1$ \quad
  \textbf{Batuhan Tosun}$^1$ \quad
  \textbf{Martin Piccolrovazzi}$^1$ \quad
  \textbf{Alexander Griessel}$^2$ \\
  \textbf{Wolfgang Kellerer}$^2$ \quad
  \textbf{Eckehard Steinbach}$^1$ \\
  $^1$Chair of Media Technology \quad $^2$Chair of Communication Networks\\
  $^1$Munich Institute of Robotics and Machine Intelligence\\
  School of Computation Information and Technology, Technical University of Munich \\
}

\begin{document}

\maketitle
\vspace{-30pt}
\begin{figure}[h!]
    \makebox[\textwidth][c]{%
        \includegraphics[width=0.8\textwidth]{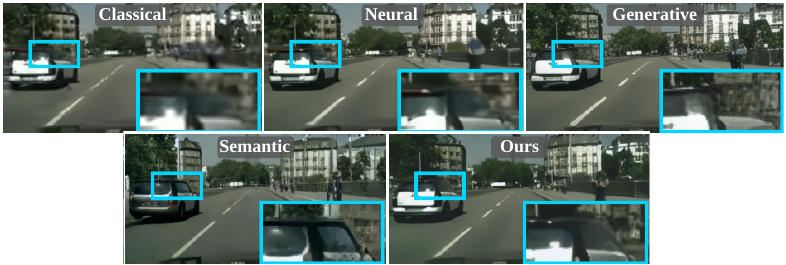}
    }
    \caption{Video reconstruction at ultra-low bitrates. Classical and neural decoders produce blurry results, generative decoders introduce distortions, and semantic reconstruction is inconsistent.}
    \label{fig:teaser}
\end{figure}

\begin{abstract}
   We study video reconstruction from ultra-low-bitrate representations, where the primary challenge shifts from encoding to decoding. In this regime, reconstruction with classical and neural codecs introduces blur, while generative and semantic approaches often struggle to jointly preserve fidelity, temporal consistency, and perceptual quality. To address these limitations, we propose a causal video diffusion model that reconstructs videos from ultra-low-bitrate semantics and highly compressed frames by jointly modeling their complementary information. We further introduce temporal-only distillation from a bidirectional teacher to enable parameter-efficient training and causal few-step inference. Through extensive quantitative, qualitative, and subjective evaluation, we show that the proposed method outperforms classical, neural, generative, and semantic baselines in ultra-low-bitrate video reconstruction.
\end{abstract}

\section{Introduction}

Video compression has traditionally focused on improving coding efficiency by optimizing the encoder through learned transforms, entropy coding, and motion modeling~\cite{gomes2025end}. With the rise of perceptual compression, emphasis has shifted toward perceptual bitrate optimization~\cite{blau2018perception,zhang2021dvc}. Driven by recent advances in generative image and video modeling~\cite{dhariwal2021diffusion,rombach2022high,peebles2023scalable,esser2024scaling}, as well as in decoder capacity, further bitrate reductions have been enabled, with transmitted representations becoming increasingly sparse and reconstruction quality no longer determined solely by encoder performance. In this regime, recovering videos from ultra-low-bitrate representations depends on a decoder that can infer missing information while preserving temporal consistency and scene structure~\cite{li2024extreme,zhang2025stablecodec,ma2025diffusion}. Classical and neural codecs, therefore, degrade substantially at such low bitrates, while generative codecs may improve perceptual quality but often introduce distortions or temporal instability. Semantic codecs instead decouple pixel-level encoding by transmitting explicit semantics and reconstructing the video with generative decoders~\cite{konuko2021ultra,grassucci2024enhancing,wan2025m3,eteke2025i2v}. However, semantics alone often fail to preserve texture fidelity. We illustrate these cases in Fig.~\ref{fig:teaser}.

To address these limitations, we propose a causal video diffusion model for reconstruction from ultra-low-bitrate representations. Our key idea is to treat ultra-low-rate video coding as a reconstruction problem under extremely sparse yet complementary inputs. We make three contributions. First, we introduce a framework that jointly models explicit semantic signals and highly compressed frames to recover both scene structure and appearance. Second, we propose a causal Temporal Adapter that enables temporally consistent autoregressive synthesis. Third, we introduce temporal-only distillation from a bidirectional teacher to a causal student, enabling parameter-efficient training and few-step inference. Quantitative, qualitative, and subjective experiments at ultra-low bitrates show that the proposed method improves perceptual quality, semantic consistency, and fidelity over classical, neural, generative, and semantic baselines. These results support a reconstruction-centric perspective for ultra-low-bitrate video coding, where the primary challenge shifts to the reconstruction process.

\section{Related Work}

While classical codecs such as VVC rely on predefined transforms~\cite{vvc}, learned compression methods optimize rate-distortion end-to-end. Early work by Ball\'e et al. introduced learned transforms that map signals to entropy-constrained latent representations, from which decoder networks reconstruct the input, typically under pixel-wise fidelity objectives~\cite{balle2016end,balle2018variational}. Video compression extends image compression through motion compensation, and neural video codecs adopt the same principle in a learned manner~\cite{chen2017deepcoder,lu2019dvc,liu2020learned,hu2021fvc,li2021deep,lu2024deep}, with DCVC further introducing deep spatiotemporal models that jointly optimize motion prediction and residual coding end-to-end~\cite{jia2025towards}. Despite these advances in encoder-side representation and temporal modeling, classical and neural codecs remain limited for reconstruction from ultra-low-bitrate representations, as shown in Fig.~\ref{fig:teaser}. This limitation motivates a shift toward decoder-focused models capable of reconstructing from extremely limited information.

Blau et al. highlighted the rate-distortion-perception trade-off, where pixel-wise fidelity can be sacrificed to achieve lower bitrates with higher perceptual quality, increasing the importance of the decoder~\cite{blau2018perception}. In this context, Agustsson et al. used a generative adversarial network (GAN) to reconstruct high-frequency information from low-bitrate representations~\cite{agustsson2019generative}. Mentzer et al. later extended this with a perceptual loss formulation~\cite{mentzer2020high}. More recently, generative latent compression methods achieved lower bitrates at high perceptual fidelity by reconstructing the latent space of a large-scale, pretrained vector-quantized variational autoencoder~\cite{jia2024generative}. Building on this, Zhang et al. extended generative latent coding to latent diffusion models~\cite{zhang2025stablecodec}, and Park et al. later introduced one-step diffusion for efficiency~\cite{park2026single}. However, these methods do not address temporal consistency.

The rate-distortion-perception trade-off has also been explored for video reconstruction. PLVC employed GANs for perceptual reconstruction~\cite{yang2022perceptual}. Du et al. later extended this with a Transformer-GAN hybrid architecture~\cite{du2024cgvc}. GLC introduced generative latent coding for videos with a latent buffer~\cite{qi2025generative}, and was subsequently extended to diffusion-based~\cite{guo2025generative} and one-step diffusion-based reconstruction~\cite{ma2025diffvc}. Wang et al. introduced explicit motion modeling with trajectory guidance for low-bitrate video reconstruction~\cite{Wang_Man_Li_Wang_Fan_Zhao_2026}. Conversely, Li et al. combined classical intra-frame coding with diffusion-based inter-frame interpolation, enabling extreme video coding (EVC) without explicit motion~\cite{li2024extreme}. However, these methods remain limited at ultra-low bitrates, as shown in Fig.~\ref{fig:teaser}, because they continue to rely solely on compressed latent representations. In contrast, our method reconstructs videos from ultra-low-bitrate semantics, shifting emphasis away from the encoder.

The family of semantic compression or communication methods, rather than relying on latent representations, reconstructs videos from ultra-low-bitrate explicit semantics~\cite{grassucci2024enhancing,xie2021deep,qin2024ai}. Konuko et al. used facial keypoints to animate guidance frames via GAN-based animation~\cite{konuko2021ultra}. To generalize beyond facial videos, later work introduced reconstruction from low-bitrate semantic maps~\cite{huang2022toward}. Generative models were further shown to be robust to semantic distortion, enabling lossy ultra-low-bitrate compression of semantic images and videos~\cite{eteke2024lossy,grassucci2025lightweight}. This direction was extended to controllable image diffusion~\cite{grassucci2024diffusion,yin2025generative} and video diffusion models~\cite{eteke2025i2v}. However, semantic-only reconstruction introduces inconsistencies as shown in Fig.~\ref{fig:teaser}. To address this, hybrid approaches combine semantic reconstruction with compression~\cite{wan2025m3,park2025transmit,eteke2025i2v}. Their performance, however, remains constrained by the underlying codec, leading to severe distortions or inconsistencies at ultra-low bitrates. Our method fills this gap by proposing a causal video diffusion model for reconstruction from ultra-low-bitrate semantics and heavily compressed frames.

\section{Methodology}
\label{sec:methodology}

\begin{figure}[!t]
 \centering
 \centering
    \includegraphics[width=0.95\linewidth]{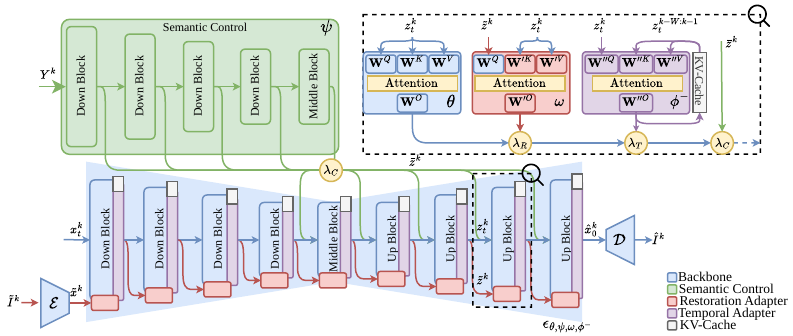}
    \caption{The proposed diffusion-based framework for ultra-low-bitrate video reconstruction. For frame $k$, the model reconstructs $\hat{I}^k$ from the diffusion latent $x_t^k$, degraded latent $\tilde{x}^k$, and sparse semantic representation $Y^k$, while leveraging causal temporal modeling within a window $z_t^{k-W:k}$.}
\label{fig:architecture}
\end{figure}

Figure \ref{fig:architecture} summarizes our causal video diffusion architecture for ultra-low-bitrate video reconstruction. Built on a latent diffusion backbone, the model incorporates Semantic Control, Restoration Adapter, and a causal Temporal Adapter, with emphasis on decoder-focused reconstruction from ultra-low-bitrate representations into temporally consistent, perceptually high-quality videos.

\subsection{Latent diffusion backbone} \label{sec:diffusion_backbone}

We start from a pretrained latent diffusion model (LDM) $\epsilon_\theta$, which performs denoising in latent space~\cite{rombach2022high}. For an input frame $I \in \mathbb{R}^{H \times W \times C}$, the LDM encoder maps the frame to a latent representation $x_0 = \mathcal{E}(I) \in \mathbb{R}^{H' \times W' \times C_e}$. Following the standard denoising diffusion probabilistic model (DDPM) formulation~\cite{ho2020denoising}, the forward process gradually perturbs $x_0$ over diffusion steps $t \in \{1,\dots,T\}$:
\begin{equation}
    \label{eq:ddpm}
    q_t(x_t \mid x_0)
    =
    \mathcal{N}\!\left(
    x_t;
    \sqrt{\bar{\alpha}_t}\,x_0,\,
    (1-\bar{\alpha}_t)\mathbf{I}
    \right),
    \qquad
    x_T \sim \mathcal{N}(0, \mathbf{I}),
\end{equation}
where $\bar{\alpha}_t$ denotes the cumulative schedule parameter. Following the variational diffusion objective, the latent denoiser $\epsilon_\theta$ is optimized via
\begin{equation}
    \label{eq:eps_loss}
    \mathcal{L}_{D}
    =
    \underset{\varepsilon,x_0,t}{\mathbb{E}}
    \left[
    \left\|
    \varepsilon - \epsilon_\theta(x_t,t)
    \right\|_2^2
    \right],
    \qquad
    \varepsilon \sim \mathcal{N}(0,\mathbf{I}),
    \qquad
    t \sim \mathcal{U}(\{1,\dots,T\}).
\end{equation}

For sampling, we adopt denoising diffusion implicit models (DDIM)~\cite{song2020denoising}, which define a non-Markovian reverse diffusion process that maps $x_T$ to $x_0$:
\begin{equation}
\label{eq:ddim}
x_{t-1}
=
\sqrt{\bar{\alpha}_{t-1}}\,x_{t \rightarrow 0}
+
\sqrt{1-\bar{\alpha}_{t-1}}\,\epsilon_\theta(x_t,t),
\qquad
x_{t \rightarrow 0}
=
\frac{x_t-\sqrt{1-\bar{\alpha}_t}\,\epsilon_\theta(x_t,t)}{\sqrt{\bar{\alpha}_t}},
\end{equation} where $x_T \sim \mathcal{N}(0, \mathbf{I})$. Finally, the LDM decoder reconstructs a frame from the denoised latent representation, $\hat{I} = \mathcal{D}(x_0)$.

Architecturally, $\epsilon_\theta$ is composed of downsampling, middle, and upsampling blocks, each containing an attention layer. Let $z_t \in \mathbb{R}^{L \times C_h}$ denote the input to one such layer. The attention operation is defined as
\begin{equation}
Q = z_t W^Q,\quad K = z_t W^K,\quad V = z_t W^V,
\end{equation}
\begin{equation}
\label{eq:attention}
\operatorname{Attention}(z_t)
=
\operatorname{softmax}\!\left(\frac{QK^\top}{\sqrt{d}}\right)V,
\end{equation}
where $W^Q, W^K, W^V \in \mathbb{R}^{C_h \times d}$ are learned attention operators~\cite{vaswani2017attention}.

Building on this formulation, our goal is to reconstruct video frames $\hat{I}^k$ from ultra-low-bitrate representations by adapting the diffusion denoising process and its attention-based architecture to preserve semantic content, perceptual fidelity, and temporal consistency.

\subsection{Semantic control} \label{sec:semantic_control}

%Explicit semantics such as facial keypoints~\cite{konuko2021ultra} or semantic maps enable extreme-low-bitrate video representation~\cite{eteke2025i2v}. 
Structured frame representations, combined with generative modeling, support ultra-low-bitrate semantic coding~\cite{eteke2024lossy}. In this setting, the semantics of the $k^{\text{th}}$ frame are represented by $Y^k$ where each element $Y^k(i,j)\in \mathbb{N}$ is a discrete class label in frame $I^k$.

When solving the DDIM reverse process in Eq.~\ref{eq:ddim} without conditioning, the model reconstructs inconsistent frames. Large-scale LDMs have been shown to enable control using semantic residuals without training from scratch~\cite{zhang2023adding}. As displayed under Semantic Control in Fig.~\ref{fig:architecture}, we utilize these semantic residuals to enable synthesis from the ultra-low-bitrate semantic maps. This approach uses a mirror of the downsampling blocks of $\epsilon_\theta$, namely $\epsilon_\psi(Y^k)$, and injects the output of each layer $\bar{z}^k$ into the upsampling blocks.

\begin{equation} \label{eq:control}
    z^k_t \leftarrow z^k_t + \lambda_C \bar{z}^k,
\end{equation} where \mbox{$\lambda_C \in [0,1]$} controls the strength of Semantic Control, and we extend the denoiser with control parameters $\psi$ to introduce $\epsilon_{\theta,\psi}(x^k_t, Y^k, t)$ shown in Fig.~\ref{fig:architecture}. Here, only the control parameters $\psi$ are trained. This model respects the scene structure conveyed by sparse $Y^k$ while still freely synthesizing texture details. This free synthesis enables reconstruction with ultra-low-bitrate semantics at the cost of fidelity.

\subsection{Restoration adapter} \label{sec:restoration}

To improve fidelity at ultra-low bitrates, we propose using heavily compressed, i.e., degraded, frames $\tilde{I}^k$ with latents $\tilde{x}^k = \mathcal{E}(\tilde{I}^k)$. At low bitrates, these latents provide low-frequency information about the frame. At each layer, the backbone produces degraded features $\tilde{z}^k$ in parallel with $z_t^k$, between which LDMs preserve similarity~\cite{eteke2025bir}. As displayed in Fig.~\ref{fig:architecture}, the Restoration Adapter extends the attention mechanism of Eq.~\ref{eq:attention} for frame restoration:
\begin{equation} \label{eq:restore}
z_t^k \leftarrow\;
\operatorname{Attention}(Q,K,V)\,W^O
+
\lambda_R\,\operatorname{Attention}(\tilde{Q},K',V')\,W'^O,
\end{equation}
where $\tilde{Q} = \tilde{z}^k W^Q$, $K' = z_t^k W'^K$, and $V' = z_t^k W'^V$. The scalar $\lambda_R \in [0,1]$ controls the strength of the restoration.

Intuitively, the degraded features query the latent space for their clean counterparts, allowing the model to restore high-frequency details guided by the heavily compressed frames. We keep $W^Q$ fixed and train only the small set of parameters $\{W'^K, W'^V, W'^O\}$ on top of the frozen backbone, which defines the parameter set $\omega$ of the extended model $\epsilon_{\theta,\psi,\omega}(x_t^k, \tilde{x}^k, Y^k, t)$ shown in Fig.~\ref{fig:architecture}.

\subsection{Temporal Adapter}
\label{sec:temporal_adapter}

While Semantic Control and Restoration Adapter enable the model to synthesize high-fidelity frames conditioned on ultra-low-bitrate video representations, both operate independently on each frame, leading to temporal inconsistencies. 

Let $z_t^{1:K} \in \mathbb{R}^{K \times L \times C_h}$ denote the activations of $K$ frames at diffusion step $t$. To apply temporal attention, we reorder tensor dimensions to $\mathbb{R}^{K \times L \times C_h } \rightarrow \mathbb{R}^{L \times K \times C_h }$ and extend the attention of Eq.~\ref{eq:attention} as shown in Fig.~\ref{fig:architecture} as
\begin{equation} \label{eq:temporal}
    z_t^{1:K} \leftarrow\; \operatorname{Attention}(Q, K, V)\,W^O + \lambda_T\,\operatorname{Attention}(Q'', K'', V'')\,W''^O,
\end{equation}
where $Q'' = z_t^{1:K} W''^Q$, $K'' = z_t^{1:K} W''^K$, and $V'' = z_t^{1:K} W''^V$ and $\lambda_T \in [0,1]$ controls the temporal contribution. The collection of all temporal-attention parameters $\{W''^Q, W''^K, W''^V, W''^O \}$ constitutes the set $\phi$ and extends the model to $\epsilon_{\theta,\omega,\psi,\phi}(x^{1:K}_t, \tilde{x}_t^{1:K}, Y^{1:K}, t)$ as seen in Fig.\ref{fig:architecture}. We apply the Temporal Adapter after the Restoration Adapter, as temporality should be accounted for in the final versions of the frames. We display our overall architecture in Fig.~\ref{fig:architecture}.

\begin{figure}[!t]
 \centering
    \includegraphics[width=1.0\linewidth]{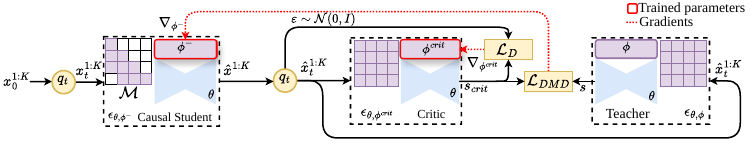}
    \caption{The temporal-only distribution matching distillation.}
\label{fig:distillation}
\end{figure}

\subsection{Causal few-step reconstruction with temporal-only distillation} \label{sec:autoregressive}

As presented in Eq.~\ref{eq:ddim}, LDMs require iterative denoising steps for reconstruction. In addition, the Temporal Adapter described above operates over a fixed number of $K$ frames using bidirectional temporal attention. Such a formulation is unsuitable for streaming and online video restoration, where frames must be reconstructed sequentially under latency constraints.

To overcome these issues, we propose temporal-only distillation, where a student model's Temporal Adapter parameters $\phi^-$ are trained using bidirectional $\epsilon_{\theta,\phi}$ as the teacher model. Notice that we drop $\psi$ and $\omega$ here, since we perform this distillation for a general temporal model independent of the ultra-low-bitrate representations described above.

Different from $\epsilon_{\theta,\phi}$, the student model, $\epsilon_{\theta,\phi^-}$, attends to a block-causal window of size $W$ by masking the attention in Eq.~\ref{eq:attention} via

\begin{equation}
    \mathcal{M}(i,j) =
    \begin{cases}
        1, & \text{if } \left\lfloor \dfrac{j-1}{W} \right\rfloor \le \left\lfloor \dfrac{i-1}{W} \right\rfloor, \\
        0, & \text{otherwise,}
    \end{cases}
    \label{eq:causal_mask}
\end{equation} which defines a block-causal attention mechanism, where bidirectional attention is allowed within each window of size $W$, while attention across windows is restricted to past blocks.

We distill the student model for causal few-step reconstruction via distribution matching distillation (DMD)~\cite{yin2024one} using the approximate gradient of the KL-Divergence between a critic and the teacher model via score distillation, skipping the teacher model's Jacobian $ \frac{\partial \epsilon_{\theta,\phi}(\hat{x}^{1:K}, t)}{\partial \hat{x}^{1:K}}$~\cite{Poole2022DreamFusionTU} as follows

\begin{equation} \label{eq:dmd}
    \nabla_{\phi^-} \mathcal{L}_{DMD} = \mathbb{E} \left[ (s_{crit}(\hat{x}^{1:K}_t, t) - s(\hat{x}^{1:K}_t, t)) \frac{\partial \hat{x}^{1:K}}{\partial \phi^-} \right], 
\end{equation} where $\hat{x}^{1:K}$ is the estimated noise-free latents of the student model as in Eq.~\ref{eq:ddim} and computed as

\begin{equation}
    \hat{x}^{1:K} =\frac{x^{1:K}_t-\sqrt{1-\bar{\alpha}_t}\epsilon_{\theta,\phi^-}(x_t,t;\mathcal{M})}{\sqrt{\bar{\alpha}_t}},
\end{equation}

$\hat{x}^{1:K}_t \sim q_t(\hat{x}^{1:K}_t|\hat{x}^{1:K})$ is the forward model in Eq.~\ref{eq:ddpm} with $t\sim\mathcal{U}(T_{min}, T_{max})$, and $s_{crit}$ and $s$ are the scores, which correspond to $-\hat{x}^{1:K}_{t\rightarrow0}$ estimated via the bidirectional critic $\epsilon_{\theta,\phi^{crit}}$ and the teacher model $\epsilon_{\theta,\phi}$ respectively~\cite{song2020score}. The optimization alternates between the student model minimizing $\mathcal{L}_{DMD}$ and the critic model minimizing $\mathcal{L}_D$ of Eq.~\ref{eq:eps_loss}. Fig.~\ref{fig:distillation} visualizes the temporal-only distillation. We further show in the appendix that our proposed temporal-only distillation significantly improves training efficiency.

Observe that once a window is processed, the corresponding activations remain fixed for subsequent windows. This enables caching of key-value (KV) pairs at the window level. We leverage this property to implement a KV-cache for efficient autoregressive inference. This finalizes the proposed causal video diffusion model for ultra-low-bitrate video reconstruction.

\begin{equation}
    \epsilon_{\theta,\omega,\psi,\phi^-}(x^{k-W:k}_t, \tilde{x}_t^{k-W:k}, Y^{k-W:k}, t).
\end{equation}

\section{Setup \& Experiments} \label{sec:experiments}

\subsection{Datasets, training, and ultra-low-bitrate representations} \label{sec:setup_info}

Our method reconstructs videos from two complementary ultra-low-bitrate inputs: (i) heavily compressed video frames, which preserve coarse appearance and low-frequency information, and (ii) heavily compressed semantic maps, which preserve scene structure and semantics. We downsample original frames by $4\times$, compress them using DCVC~\cite{jia2025towards}, and pair them with contour-based lossy semantic map compression~\cite{eteke2024lossy}. Our focus is on reconstruction quality from these representations rather than encoder design.

We train Semantic Control (Sec.~\ref{sec:semantic_control}) on datasets with semantic annotations. We use Cityscapes~\cite{Cordts2016Cityscapes}, processed at $256 \times 512$ and 10 fps, where full-sequence semantic maps are obtained using SAM2~\cite{ravi2024sam}. To complement urban scenes with different dynamics, we additionally generate YCB-Sim using BlenderProc~\cite{Denninger2023} and the YCB object set~\cite{Calli2015YCB}. YCB-Sim dataset contains 10,000 videos with semantic maps at $512 \times 512$ and 10 fps, featuring randomized textures, lighting, object interactions, and non-rigid motion.

We train the Restoration Adapter (Sec.~\ref{sec:restoration}) on high-quality image datasets~\cite{div2k,gu2019div8k,lim2017enhanced,wang2018recovering} using a standard degradation model~\cite{wang2021real}. We perform the temporal-only distillation (Sec.~\ref{sec:autoregressive}) independently as a text-to-video task. In this context, we use OpenVid-1M, processed at $512 \times 512$ and 10 fps to provide diverse videos with descriptions~\cite{nan2024openvid}.

We evaluate the full model on the validation sets of Cityscapes and YCB-Sim, both containing 500 videos. Additional information regarding YCB-Sim dataset curation, training, implementation details, and lossy semantic compression is provided in the appendix.

\subsection{Metrics}
\label{sec:metrics}

% Finally, we use CLIP-IQA as a reference-free quality metric~\cite{wang2023exploring}
% , and stronger reference-free scores do not always align with reference-based perceptual metrics

We evaluate our approach using both pixel-level and perceptual quality metrics, including reference-based and reference-free measures. For pixel-level fidelity, we report PSNR. For perceptual quality, we use LPIPS~\cite{zhang2018unreasonable} and DISTS~\cite{ding2020image}. To assess semantic consistency, we compute the mean Intersection over Union (mIoU) between the outputs of a semantic segmentation model applied to synthesized and ground-truth frames. This semantic segmentation model comprises YOLO~\cite{Jocher_Ultralytics_YOLO_2023} for object detection and SAM2~\cite{ravi2024sam} for mask extraction. We primarily report results in a rate-distortion setting, where bitrate is reported in terms of bits per pixel (bpp) and controlled jointly through semantic map compression and neural frame compression. We provide more details in the appendix.

We note that conventional rate-distortion analysis only partially captures performance in generative settings. Unlike classical codecs, the final output quality is not determined solely by the distortion induced by lossy compressed representations, since the generative model can recover plausible content beyond what is explicitly preserved. Moreover, improvements in perceptual quality do not necessarily correspond to higher pixel-level fidelity, nor does higher pixel-level fidelity imply better perceptual quality. We therefore further validate our method with a subjective study.

\subsection{Baselines} \label{sec:baselines}

As our goal is to reconstruct videos at ultra-low bitrates, the proposed method is closely related to video compression. We therefore compare against a range of conventional, neural, and generative video compression methods. As a conventional codec, we use VVC~\cite{vvc}. As a neural codec, we evaluate DCVC~\cite{jia2025towards}. As generative compression baselines, we use Extreme Video Coding (EVC)~\cite{li2024extreme} and the state-of-the-art Generative Latent Codec (GLC)~\cite{qi2025generative}. In addition, we compare with I2V-SC~\cite{eteke2025i2v}, an image-to-video semantic communication framework that uses the same diffusion backbone and a temporal module with a fixed length of $K = 16$ frames. To evaluate causal generation at $K = 30$, we use the distilled Temporal Adapter within I2V-SC. We included EVC only in Cityscapes, as the publicly available model was trained only on it. Together, these baselines cover classical, neural, generative, and ultra-low-bitrate paradigms. Fig.~\ref{fig:rd} displays the rate-distortion curves.

\subsection{Ablation Studies} \label{sec:ablation}

We first examine the three weighting parameters that control the relative contribution of each module: $\lambda_C$ for Semantic Control in Eq.~\ref{eq:control}, $\lambda_R$ for the Restoration Adapter in Eq.~\ref{eq:restore}, and $\lambda_T$ for the Temporal Adapter in Eq.~\ref{eq:temporal}, as described in Sec.~\ref{sec:methodology} and illustrated in Fig.~\ref{fig:architecture}. For both datasets, we selected 25 videos and performed a grid search over these three parameters, sweeping each within $[0,1]$ in increments of $0.1$, resulting in a total of $11^3\times25=33,275$ evaluated configurations. The further evaluation results exclude these 25 videos. This analysis demonstrates how each module contributes to successful synthesis and how its interactions influence overall performance. Fig.~\ref{fig:weights} presents the effect of each weight. We fix the value of a weight and report the mean and $99\%$ confidence interval of the metrics across a sweep of the remaining two weights. We include the Consistency metric to evaluate $\lambda_T$. The Consistency metric is the cosine similarity of CLIP~\cite{Radford2021LearningTV} features between consecutive frames. We additionally evaluate the rate-distortion performance at ultra-low-bitrates when the Semantic Control and Restoration Adapter are disabled, i.e., $\lambda_C=0$ and $\lambda_R=0$, respectively.

\subsection{Subjective Evaluation} \label{sec:subjective}

Reference-based metrics, particularly pixel-level measures such as PSNR, do not fully capture the perceptual behavior of generative models, especially under the perception-distortion trade-off~\cite{blau2018perception}. We therefore complement quantitative results with an ethics committee-approved subjective user study under informed consent. Participants were shown two anonymized reconstructions of the same video in a randomized, side-by-side, forced-choice setting, in which our method was compared with competing baselines. The participants were asked to select the preferred result based on overall visual quality. Full experimental details are provided in the appendix.

\section{Results \& Discussion} \label{sec:results}

\begin{figure}[!t]
 \centering
    \includegraphics[width=0.75\columnwidth]{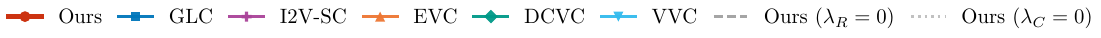}
    \vspace{2mm}
    \begin{subfigure}[t]{\textwidth}
        \centering
        \includegraphics[width=\linewidth]{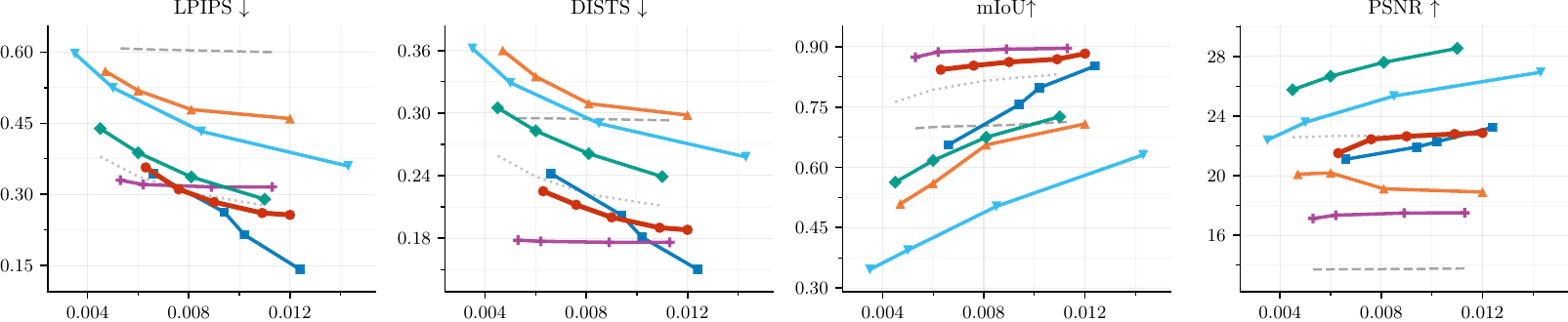}
        \caption{Cityscapes}
        \label{fig:large}
    \end{subfigure}

    \begin{subfigure}[t]{\textwidth}
        \centering
        \includegraphics[width=\linewidth]{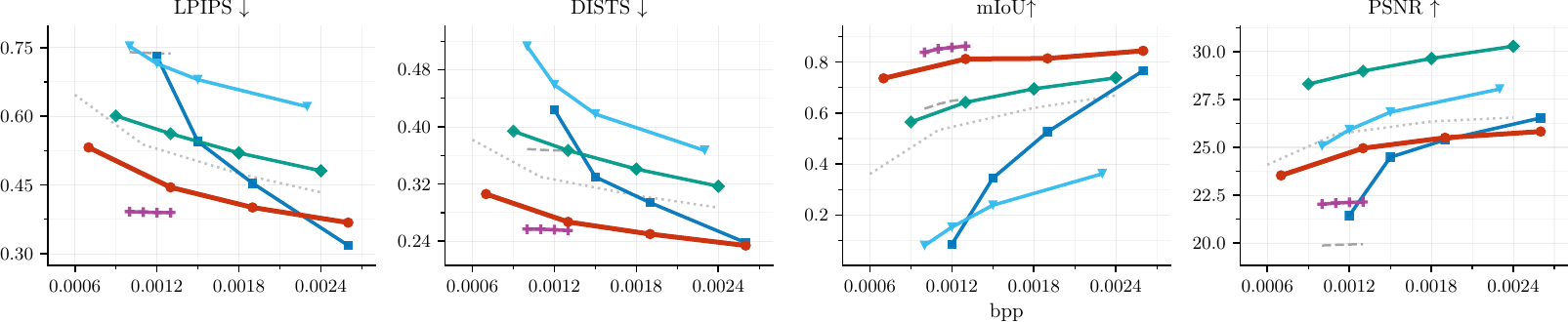}
        \caption{YCB-Sim}
        \label{fig:small}
    \end{subfigure}
\caption{Rate-distortion curves.}
\label{fig:rd}
\end{figure}

\subsection{Rate-distortion results} \label{sec:quant}

Fig.~\ref{fig:rd} reports the rate-distortion curves of the proposed method, the baselines introduced in Sec.~\ref{sec:baselines}, and ablated variants that disable Semantic Control ($\lambda_C = 0$) or the Restoration Adapter ($\lambda_R = 0$), as discussed in Sec.~\ref{sec:ablation}.

Across both datasets, our method provides the strongest overall rate-distortion-perception trade-off at ultra-low bitrates when jointly considering pixel fidelity (PSNR), perceptual quality (LPIPS, DISTS), and semantic consistency (mIoU). The conventional codec VVC and the neural codec DCVC achieve the highest PSNR values. However, perceptual and semantic metrics indicate clear degradation in this low-bitrate regime. Our method consistently outperforms the state-of-the-art generative latent compression baseline GLC at ultra-low bitrates across the reported metrics. The semantic baseline I2V-SC achieves competitive, and in some cases stronger, performance. However, as further examined in the qualitative and subjective evaluations, these scores do not fully align with perceived visual quality. A similar trend is observed for GLC at higher bitrates, where competitive quantitative results do not reflect the poorer visual quality. To support our argument, we conduct subjective comparisons at target bitrates of $0.012$ bpp for Cityscapes and $0.0019$ bpp for YCB-Sim, at which the quantitative performance gap between methods begins to tighten.

Disabling the Restoration Adapter ($\lambda_R = 0$) results in consistent losses in PSNR, LPIPS, and DISTS, highlighting its role in recovering fidelity from highly compressed frames. Disabling Semantic Control ($\lambda_C = 0$) reduces mIoU, confirming that explicit semantic guidance is important for preserving scene structure and semantics. Despite these degradations, both ablated variants remain competitive, indicating that each component provides complementary gains.

\begin{figure}[t]
    \centering
    \includegraphics[width=0.6\columnwidth]{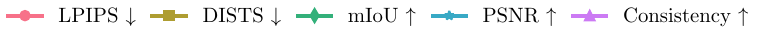}
    \vspace{2mm}
    \subfloat[Cityscapes]{%
        \includegraphics[width=0.5\linewidth]{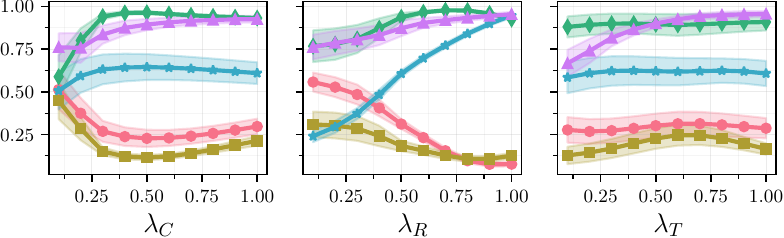}%
    }
    \subfloat[YCB-Sim]{%
        \includegraphics[width=0.5\linewidth]{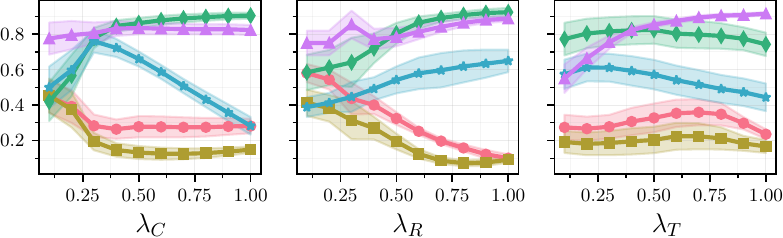}%
    }
\caption{Effect of module weights $\lambda_C$, $\lambda_R$, and $\lambda_T$. Normalized metric values show that Semantic Control ($\lambda_C$) improves semantic consistency, the Restoration Adapter ($\lambda_R$) improves fidelity and perceptual quality, and the Temporal Adapter ($\lambda_T$) improves temporal consistency.}
\label{fig:weights}
\end{figure}

\subsection{Effect of $\lambda_C$, $\lambda_R$, and $\lambda_T$} \label{sec:hyperparameters}

Fig.~\ref{fig:weights} shows the effect of the three weighting parameters introduced in Sec.~\ref{sec:methodology}: the Semantic Control weight $\lambda_C$, the Restoration Adapter weight $\lambda_R$, and the Temporal Adapter weight $\lambda_T$. Increasing $\lambda_C$ up to $0.4$ consistently improves semantic preservation and perceptual quality, after which gains saturate. In contrast, $\lambda_T$ has little effect on PSNR, but substantially improves perceptual quality and temporal consistency, with the best results observed at $\lambda_T=1.0$, highlighting its importance for coherent autoregressive synthesis. Increasing $\lambda_R$ improves both fidelity, as measured by PSNR, and perceptual quality, as measured by LPIPS and DISTS. Based on these observations, we use $\lambda_C=0.4$, $\lambda_T=1.0$, and $\lambda_R=0.8$ for all experiments. Overall, these findings indicate that the three components provide complementary benefits.

\begin{wrapfigure}[12]{l}{0.49\textwidth}
    \centering
    \vspace{-8pt}
    
    \begin{minipage}[t]{0.49\linewidth}
        \centering
        \includegraphics[width=\linewidth]{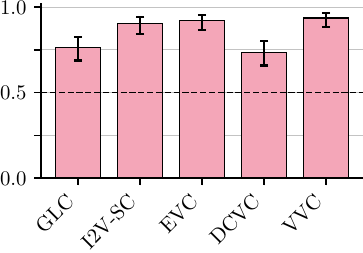}
        \caption*{(a) Cityscapes \\ ($0.012$ bpp)}
        \label{fig:preference_cityscapes}
    \end{minipage}\hfill
    \begin{minipage}[t]{0.49\linewidth}
        \centering
        \includegraphics[width=\linewidth]{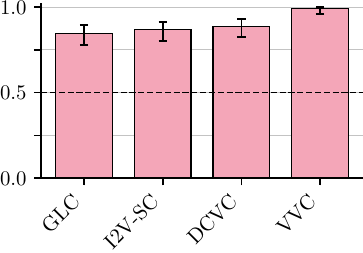}
        \caption*{(b) YCB-Sim \\ ($0.0019$ bpp)}
        \label{fig:preference_ycb}
    \end{minipage}

    \caption{Subjective preference of our method.}
    \label{fig:preference}
    \vspace{-10pt}
\end{wrapfigure}

\subsection{Subjective Evaluation}

Objective metrics do not fully capture perceptual quality, particularly for generative methods. In our setting, this mismatch is especially evident for I2V-SC and GLC (Sec.~\ref{sec:quant}). We therefore conducted a user study ($n=24$). Fig.~\ref{fig:preference} reports pairwise preference ratios with 95\% Wilson confidence intervals against each baseline. Across both datasets, our method was consistently preferred over all competing approaches, with all comparisons statistically significant after Holm correction ($p<0.05$). These results support the qualitative comparisons and further indicate that conventional objective metrics only partially reflect perceived visual quality in the ultra-low-bitrate regime.

\begin{figure*}[!t]
    \centering
    \subfloat[Cityscapes ($0.009$ bpp)]{%
        \includegraphics[width=1.0\linewidth]{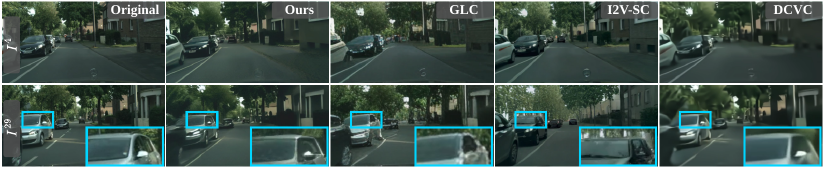}%
    }
    
    \subfloat[YCB-Sim ($0.0019$ bpp)]{%
        \includegraphics[width=1.0\linewidth]{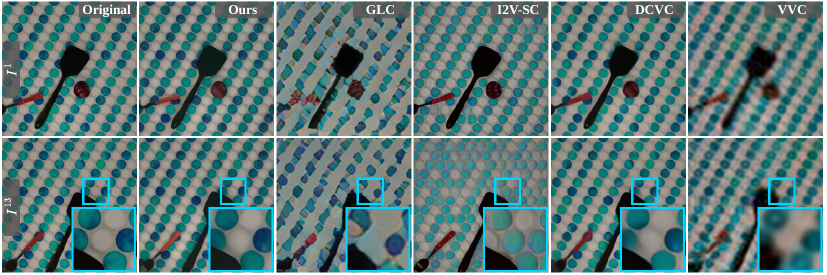}%
    }
    \caption{Qualitative results at ultra-low bitrates. VVC and DCVC appear blurry, GLC introduces visual distortions, and I2V-SC exhibits inconsistencies.}
    \label{fig:qual}
\end{figure*}

\begin{figure*}[!ht]
    \centering
    \subfloat[Ground truth]{%
        \includegraphics[width=0.245\linewidth]{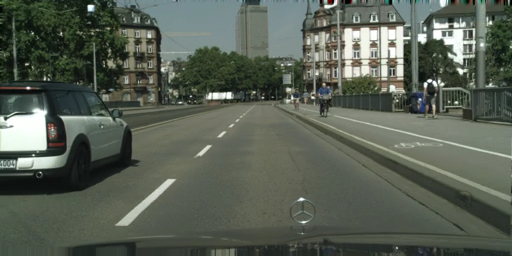}%
    }\hspace{2pt}\subfloat[Ours]{%
        \includegraphics[width=0.245\linewidth]{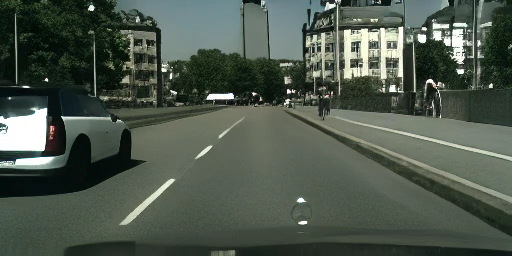}%
    }\hspace{2pt}\subfloat[Ours ($\lambda_C=0$)]{%
        \includegraphics[width=0.245\linewidth]{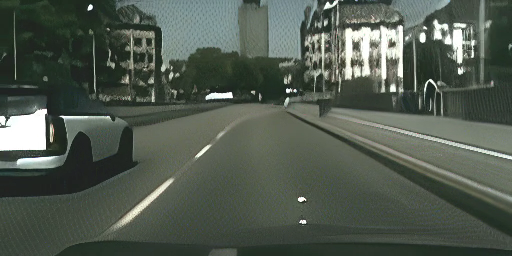}%
    }\hspace{2pt}\subfloat[Ours ($\lambda_R=0$)]{%
        \includegraphics[width=0.245\linewidth]{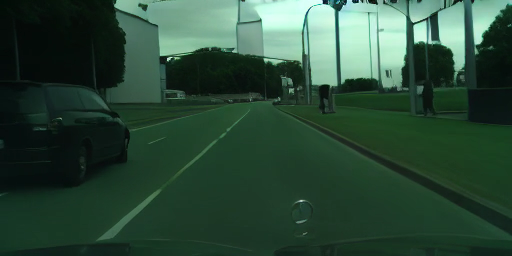}%
    }
    \caption{Qualitative ablation results. Removing Semantic Control ($\lambda_C = 0$) reduces semantic consistency, while removing the Restoration Adapter ($\lambda_R = 0$) lowers perceptual fidelity.}
    \label{fig:abl-qual}
\end{figure*}

\subsection{Qualitative Results}

Fig.~\ref{fig:qual} presents qualitative results on both datasets, comparing our approach with the baselines. The proposed method yields visually stronger reconstructions than the classical and neural codecs, better preserves consistency than the semantic baseline I2V-SC, and avoids the distortions observed in the generative latent compression baseline GLC. These comparisons support the quantitative results, the identified limitations of objective metrics, and the outcomes of the subjective study. Fig.~\ref{fig:abl-qual} further illustrates the ablation results: Semantic Control preserves semantic consistency, while the Restoration Adapter improves visual fidelity. Additional qualitative examples are provided in the appendix. We also include a supplementary website with videos at multiple bitrates.

\section{Conclusion} \label{sec:conclusion}

We presented a causal video diffusion framework for reconstruction from ultra-low-bitrate video representations, positioning the decoder as a key enabler of ultra-low-bitrate operation. The proposed method jointly models semantics and highly compressed frames within a causal diffusion-based formulation, enabling temporally consistent reconstruction with strong perceptual and semantic fidelity at ultra-low bitrates. Through temporal-only distillation, we further enabled parameter-efficient training and few-step causal synthesis suitable for streaming and as a step toward real-time reconstruction. Across quantitative and qualitative evaluations, our approach outperformed classical, neural, and generative baselines in the ultra-low-bitrate regime. We also highlighted the limitations of conventional quality metrics and demonstrated through subjective evaluations that our method is significantly preferred, even in regimes where objective metrics suggested competitive performance. These findings support a generative reconstruction-centric perspective for ultra-low-bitrate video applications. Nevertheless, our method can introduce oversmoothing and texture-sticking artifacts due to ambiguities in degraded frames and high-level semantics. Furthermore, despite using few-step diffusion, the method is not yet real-time. Future work will focus on reducing latency, improving output resolution, and strengthening temporal consistency.

\small{
\bibliographystyle{plainnat}
\bibliography{main.bib}
}

\newpage
\appendix

\section{More qualitative results} \label{sec:more_qual}
\vspace{-15pt}
\begin{figure*}[!h]
    \centering
    \subfloat{%
        \includegraphics[width=0.92\linewidth]{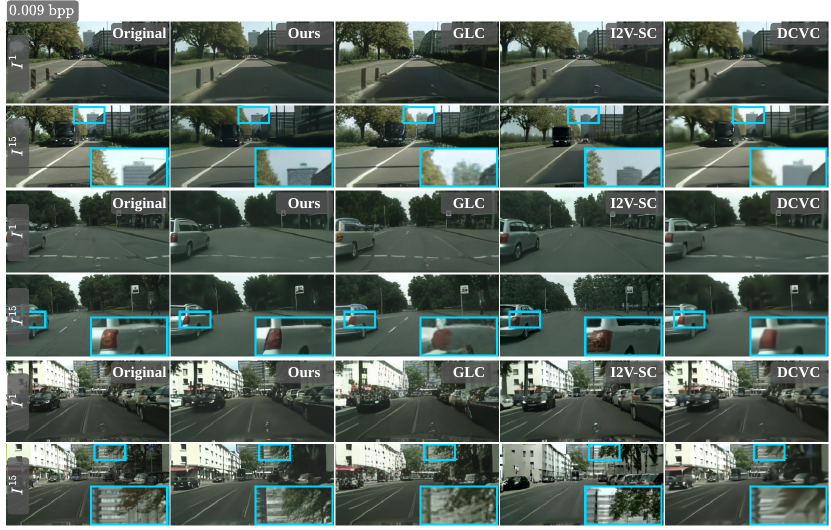}%
    }\\
     \subfloat{%
        \includegraphics[width=0.92\linewidth]{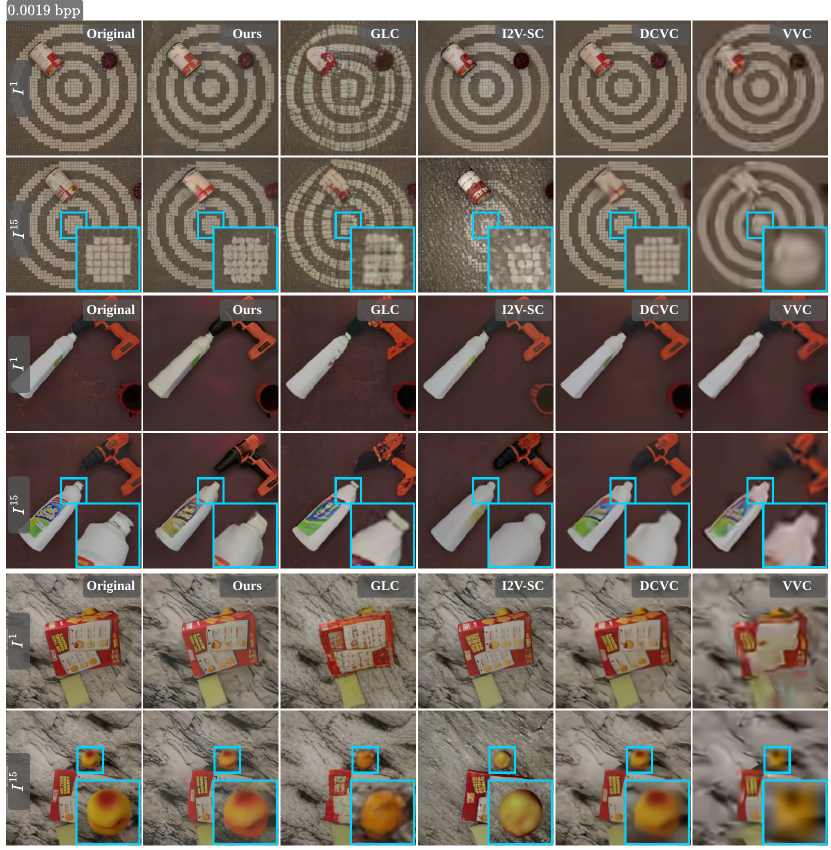}%
    }
    \caption{Qualitative results at fixed target bitrates.}
    \label{fig:fix_rates_qual}
\end{figure*}

\begin{figure*}[!t]
    \centering
    \subfloat{%
        \includegraphics[width=0.88\linewidth]{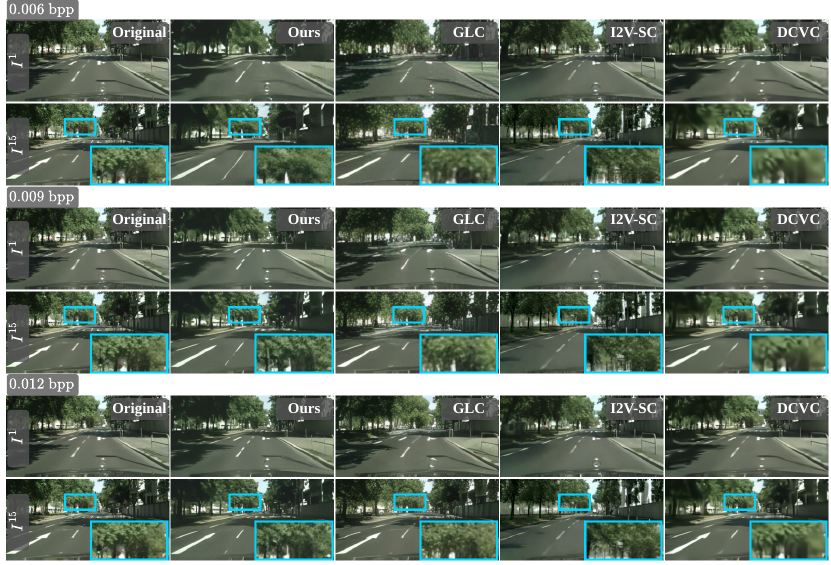}%
    }\\
     \subfloat{%
        \includegraphics[width=0.88\linewidth]{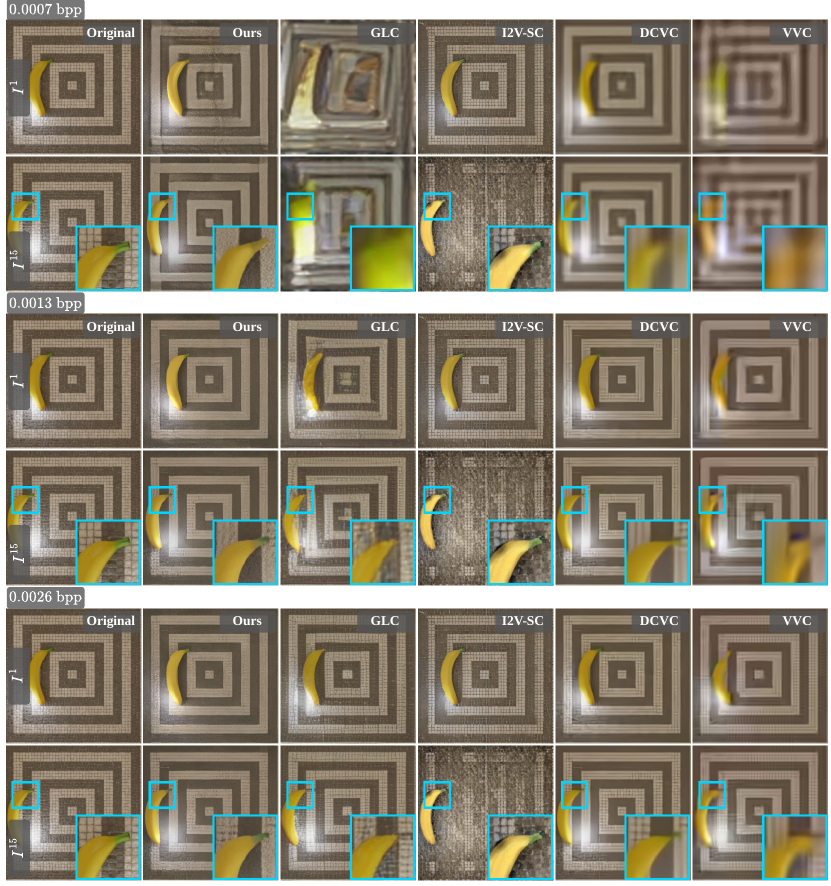}%
    }
    \caption{Qualitative results at varying target bitrates.}
    \label{fig:rates_qual}
\end{figure*}

\begin{figure}[!t]
\centering
\includegraphics[width=0.75\linewidth]{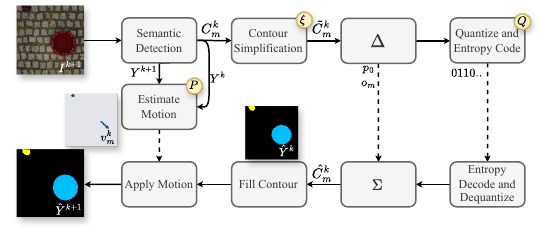}
\caption{Semantic video coding pipeline. Contours extracted from semantic object masks are simplified with a tolerance $\xi$, then differentially encoded, quantized to $Q$ symbols, and entropy-coded. For P-frames, selected via the I-frame period $P$, only semantic motion is transmitted.}
\label{fig:semantic_compression}
\end{figure}

\section{Semantic video coding}
\label{sec:semantic_video_coding}

We utilize an extended version of a semantic IP-Frame compression method~\cite{eteke2024lossy}. For an I-frame at video timestep $k$, i.e., when \mbox{$k \bmod P = 0$}, the semantic map $Y^{k}$ is decomposed into disjoint instance contours by identifying class boundaries and separating non-connected regions. We represent each object label $o$ in the local pixel neighborhood $\mathcal{N}$ of $Y^{k}$ as an ordered set of 2D contour points:

\begin{equation}
M^{k}=\{(i,j)\mid Y^{k}(i,j)=o ~\land ~\exists (i',j')\!\in\!\mathcal{N}:Y^{k}(i', j')\neq o\}\text{.}
\end{equation}

The set $M^{k}$ is then split into disjoint connected components $C_m^k$ of instances $o_m$ such that

\begin{equation}
M^{k} = \bigcup_{m} C_m^{k}\text{.}
\end{equation}
Finally, every instance is represented as a set of coordinates, i.e., $C_m^{k} := \{\, p_{m0},~p_{m1},\dots~\mid~p_{mi} \in \mathbb{R}^2\}$, by tracing the boundary points in a clockwise order.

We used line simplification with tolerance $\xi$ to create the reduced set of contours $\tilde{C}_m^{k}$ that represent the same shape  with fewer contour points. This contour extraction and simplification is implemented in OpenCV~\cite{bradski2000opencv} using border-following and Douglas-Peucker algorithms. The contour points are then differentially encoded, where the first point of each instance $p_{m0}$ and the instance identifier, i.e., the class label, $o_m$, are stored explicitly, and all remaining points are represented as differentials 
\begin{equation}
    \Delta \tilde{C}_m^k = \{p_{m1} - p_{m0}, ~p_{m2} - p_{m1} \dots\}  .
\end{equation}

The differentials $\Delta \tilde{C}_m^k$ are quantized uniformly using $Q$ symbols and entropy coded via arithmetic coding, yielding a compact representation of the entire contour. At the receiver, inverse quantization and cumulative summation reconstruct the contour $\hat{C}_m^{k}$, which is finally rasterized into a dense semantic map to reconstruct the semantics as

\begin{equation}
\hat{Y}^k_m(i,j)
= o_m \,\mathbbm{1}\!\big[(i,j) \in \mathcal{R}(\hat{C}^k_m)\big].
\end{equation}

For P-frames, we reuse the previously decoded semantic map $\hat{Y}^k$ and store only instance-wise motion $v_m$ and label $o_m$. We estimate a rigid translation vector $v_m^k\in\mathbb{R}^2$ for each instance $m$ as the shift in the centroid
\begin{equation}
v_m^k=\Big\lceil \mu_m^{k+1}-\mu_m^{k}\Big\rceil,\quad
\mu_m^{k}=\frac{\sum_{i,j}(i,j)\,\mathbbm{1}[Y^{k}(i,j)=o_m]}{\sum_{i,j}\mathbbm{1}[Y^{k}(i,j)=o_m]} .
\end{equation}
At the receiver, contour instances in $\hat{Y}^k$ are shifted to yield the next semantic map $\hat{Y}^{k+1}$ such that
\begin{equation}
\hat{Y}^{k+1}_m(i,j)
= \hat{Y}^{k}_m\!\left( \lfloor i - v_m^k(1) \rfloor,\, \lfloor j - v_m^k(2) \rfloor \right),
\end{equation}

This approach enables us to represent videos of semantic maps at ultra-low bitrates. Fig.~\ref{fig:semantic_compression} illustrates the described semantic video compression method. In this work, we proposed a method that utilizes these ultra-low-bitrate semantics for causal video reconstruction via the Semantic Control. We discussed that semantics-only reconstruction does not yield high-fidelity results; for example, a car's direction is not represented in the semantics in an urban video scenario.

\begin{figure}[t]
    \centering
    \includegraphics[width=0.95\columnwidth]{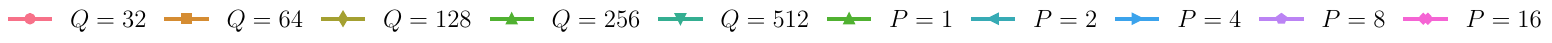}
    \vspace{2mm}
    \subfloat[Cityscapes]{%
        \includegraphics[width=0.5\linewidth]{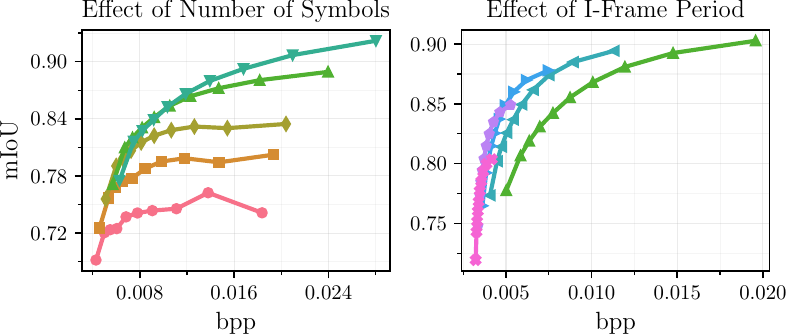}%
        \label{fig:rd_citycscapes}
    }
    \subfloat[YCB-Sim]{%
        \includegraphics[width=0.5\linewidth]{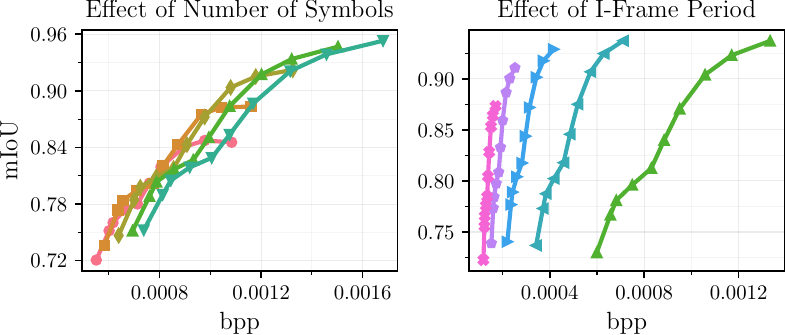}%
        \label{fig:rd_ycb}
    }
    \caption{The semantic rate-distortion curves. We investigate the effect of the number of quantization symbols $Q$ and the I-Frame period $P$, where we control the bitrate on the x-axis by setting the contour simplification tolerance $\xi \in [4, 24]$.}
    \label{fig:semantic_rd}
\end{figure}

\subsection{Evaluation}

The semantic video codec, described in Sec.~\ref{sec:semantic_video_coding}, is governed by three parameters that control the bitrate and semantic distortion: the contour simplification tolerance $\xi$, the I-Frame period $P$, and the number of quantization symbols $Q$.We analyze the effect of each parameter on the semantic fidelity, measured by the mean Intersection over Union (mIoU), and the semantic distortion across both datasets. The results are summarized in Fig.~\ref{fig:semantic_rd}. Throughout the experiments, we sweep $\xi \in [4, 24]$, which serves as the primary bitrate–control parameter for each choice of $P$ and $Q$, allowing us to study their interactions.

\subsubsection{Effect of $Q$ and $\xi$}

Parameters $Q$ and $\xi$ jointly determine the bitrate of I-Frames. The resulting rate-distortion curves, shown in Fig.~\ref{fig:rd}, indicate that $Q = 256$ offers the best trade-off between bitrate and semantic fidelity. In both datasets, for example, the configuration $Q = 256$ and $\xi = 6$ achieves $\text{mIoU} \geq 0.85$ while enabling ultra-low-bitrate operation: approximately $0.011$ bpp on Cityscapes and $0.001$ bpp on YCB-Sim. Throughout our work, we fixed $Q=256$ and varied $\xi$ depending on the bitrate level.

\subsubsection{Effect of $P$ and $\xi$}
$P$ determines the number of P-frames following each I-frame. Fixing $Q = 256$ and sweeping $\xi$ as before, we obtain the curves presented in Fig.~\ref{fig:semantic_rd}. Larger values of $P$ introduce greater distortion, while smaller values reduce coding efficiency by requiring more frequent I-Frames. Setting, for example, $Q=256$, $\xi=6$, and $P=4$ allows the semantics to be represented at ultra-low bitrates of $0.006$ bpp and $0.0007$ bpp for Cityscapes and YCB-Sim, respectively. Cityscapes contain dense, large-scale urban scenes with complex object boundaries, which naturally require more bits to encode. In contrast, YCB-Sim features comparatively sparse tabletop arrangements with fewer, well-separated semantic instances, allowing the contour-based representation to reach substantially lower bitrates.

\subsubsection{Bitrate settings} \label{sec:rate_settings}

In our work, we jointly model ultra-low-bitrate semantics and heavily compressed ultra-low-bitrate frames. Hence, the bitrate is determined by the parameters $P$ and $\xi$ with $Q=256$ and the video compression rate. As we discussed in Sec.~\ref {sec:setup_info}, we utilized DCVC for heavy compression and utilized downsampling to reach ultra-low bitrates~\cite{jia2025towards}. In Tab.~\ref{tab:rate_params}, we present the settings we used for both datasets to achieve the reported bitrates.

\begin{table}[h!]
\centering
\caption{Bitrate control parameters: $P$ controls the semantic I-Frame period, $\xi$ controls the contour tolerance, $Q_V$ controls the DCVC quantization, and $\downarrow$ controls the downsampling rate.}
\scriptsize{
    \begin{tabular}{@{}ccccc|cccc|cccc|cccc@{}}
    \toprule
     & $P$ & $\xi$ & $Q_{V}$ & $\downarrow$ & $P$ & $\xi$ & $Q_{V}$ & $\downarrow$ & $P$ & $\xi$ & $Q_{V}$ & $\downarrow$ & $P$ & $\xi$ & $Q_{V}$ & $\downarrow$ \\ \midrule
    Cityscapes & 16 & 12 & 25 & 4 & 16 & 12 & 15 & 2 & 16 & 12 & 20 & 2 & 4 & 12 & 25 & 2 \\
    bpp & \multicolumn{4}{c|}{0.0063} & \multicolumn{4}{c|}{0.0076} & \multicolumn{4}{c|}{0.009} & \multicolumn{4}{c}{0.0109} \\ \midrule
    YCB & 16 & 10 & 30 & 8 & 8 & 10 & 25 & 4 & 16 & 10 & 15 & 2 & 8 & 10 & 29 & 2 \\
    bpp & \multicolumn{4}{c|}{0.0007} & \multicolumn{4}{c|}{0.0013} & \multicolumn{4}{c|}{0.0019} & \multicolumn{4}{c}{0.0026} \\ \bottomrule
    \end{tabular}
}
\label{tab:rate_params}
\end{table}

\begin{figure}[t]
    \centering
    \includegraphics[width=0.83\linewidth]{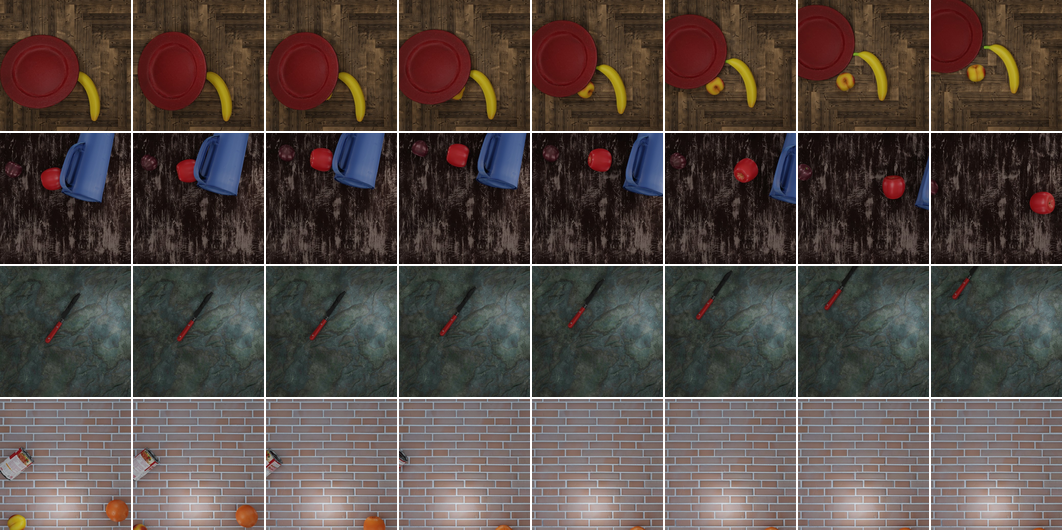}%
    \caption{Example videos from the YCB-Sim dataset.}
    \label{fig:ycb}
\end{figure}

\section{YCB-Sim dataset} \label{sec:ycb_sim}

\begin{algorithm}[h!]
\caption{Synthetic Scene Generation}
\label{alg:data_gen}
\footnotesize
\begin{algorithmic}[1]
\Require scene id $i$, base seed $s$, asset/config paths
\State Seed RNG with $s+i$; initialize BlenderProc~\cite{Denninger2023}
\State Set camera intrinsics $K=(f_x,f_y,c_x,c_y)=(710,710,256,256)$, image size $512{\times}512$
\State Set renderer: noise threshold $0.01$, max samples $4096$, JPEG quality $90$
\State Set simulation timing: $10$ FPS, force start $2$s, render window $[2,5]$s
\State Create desk plane of size $0.25$m and a fixed top-down camera at height $0.65$m
\State Load YCB object meshes~\cite{Calli2015YCB} and CCTexture
\State Sample $n\sim \mathcal\{1,2,3\}$ object categories without replacement
\ForAll{sampled objects}
    \State Enable rigid body dynamics, assign class id
    \State Sample material roughness from $U(0.5,1.0)$ and specular from $U(0,0.5)$
\EndFor
\State Sample one CCTexture for the desk/background;
\State Create an area light above the desk at height $0.325$m
\Statex \hspace{1em}sample emission from $U(0.5,1.0)$
\State Create a point light
\Statex \hspace{1em}sample energy from $U(35,70)$
\Statex \hspace{1em}sample position on hemisphere above the table 
\Statex \hspace{2em} radius $\sim\mathcal{U}[0.1625,0.325]$m and elevation $\sim \mathcal{U}[10^\circ,80^\circ]$
\For{$k=1$ to $2N$} \Comment{$N=4$}
    \State Add a wind field on the desk perimeter, oriented toward the desk center
    \State Field strength $\sim U(1,7)$
\EndFor
\For{$k=1$ to $2N$}
    \State Add a point-force field at a random desk location
    \State falloff $\sim U(0,2)$, strength $\sim U(1,7)$
\EndFor
\State Sample non-overlapping initial object poses over the desk
\State Run physics simulation for $5$s; at $3$s activate first $N$ forces, at $4$s activate second $N$ forces
\State Render frames from $2$ to $5$s and output RGB and semantic masks
\end{algorithmic}
\end{algorithm}

One of the datasets used to evaluate our method is the simulated YCB-Sim dataset. We generated this dataset using BlenderProc2~\cite{Denninger2023} with publicly available YCB object models~\cite{Calli2015YCB} and CC material textures\footnote{\url{https://dlr-rm.github.io/BlenderProc/examples/datasets/scenenet_with_cctextures/README.html}}, both licensed under CC-BY 4.0. We created this benchmark because densely annotated semantic video datasets remain relatively limited, particularly for object-centric scenes with rich physical interactions outside urban driving environments. Since our ultra-low-bitrate video reconstruction framework explicitly leverages ultra-low-bitrate semantic maps, we required sequences with accurate frame-wise labels with complex object interactions.

To generate the data, we randomly sample YCB objects, place them above a tabletop scene with randomized poses, materials, lighting, and backgrounds, and simulate two temporally separated phases of external forces to induce complex object interactions and motions. We provide the full generation procedure and parameters in Alg.~\ref{alg:data_gen}, with example frames shown in Fig.~\ref{fig:ycb}.

\section{Implementation details} \label{sec:implementation}

In this section, we describe the implementation details of our proposed method presented in Sec.~\ref{sec:methodology}, along with the corresponding training procedures and key hyperparameters. All the training discussed below is performed on $4\times$ NVIDIA RTX Pro 6000 Blackwell GPUs using the AdamW optimizer with a weight decay of $10^{-2}$ and running averages $\beta_1$ and $\beta_2$ of $0.9$ and $0.999$, respectively.

\subsection{Diffusion backbone}

As described in Sec.~\ref{sec:diffusion_backbone}, our approach uses a pretrained and frozen latent diffusion backbone. We use Stable Diffusion~\cite{podell2023sdxl} as the underlying denoiser, operating with an empty text prompt since no textual conditioning is required. This backbone provides a strong image prior for high-quality synthesis, and keeping it fixed stabilizes training across all modules while allowing each adapter to specialize efficiently.

\subsection{Semantic control} \label{sec:semantic_control_implementation}

To condition the synthesis on the transmitted semantic maps, we use a residual-based control mechanism as described in Sec.~\ref{sec:semantic_control}. For this purpose, we employ ControlNet~\cite{zhang2023adding} and train separate sets of control parameters $\psi$ for Cityscapes and YCB-Sim. This allows us to switch between datasets without retraining the full model. We train the control module on labeled frames from each dataset using a batch size of 16 (effective batch size $16\times 4 = 64$), a learning rate of $2\times 10^{-6}$, and gradient clipping of $1$ for $10^5$ iterations. The objective is the diffusion loss in Eq.~\ref{eq:eps_loss}, computed using the backbone extended with the Semantic Control, $\epsilon_{\theta,\psi}(x_t, Y_t, t)$. This module plays a crucial role in preserving scene structure and ensuring that synthesized frames retain semantic fidelity. We presented the ablation studies in Sec.~\ref{sec:ablation}. We provided details about YCB-Sim in the previous section. We utilized the Cityscapes datasets in accordance with its license\footnote{\url{https://www.cityscapes-dataset.com/license/}} for research purposes only.

\subsection{Restoration adapter}

We discussed in Sec.~\ref{sec:restoration} that we incorporate the ultra-low bitrate, low-resolution, compressed frames into the reconstruction for fidelity. To that extent, we employ the BIR-Adapter~\cite{eteke2025bir}. Restoration Adapter learns restoration-specific attention parameters $\omega$ that guide the diffusion backbone toward the clean latent features corresponding to the degraded inputs. We train the adapter using a combination of DIV2K~\cite{div2k}, DIV8K~\cite{gu2019div8k}, OutdoorSceneTrain~\cite{wang2018recovering}, and Flickr2K~\cite{lim2017enhanced}, applying a widely used degradation model that includes $4\times$ downsampling, noise, blur, and compression artifacts~\cite{wang2021real}. Training uses random $512\times512$ patches, a batch size of 16 (effective $16\times 4 = 64$), a learning rate of $10^{-5}$, and $10^5$ iterations while minimizing the diffusion loss in Eq.~\ref{eq:eps_loss} evaluated with the restoration-extended backbone $\epsilon_{\theta,\omega}(x_t, \tilde{x}_t, t)$. This adapter proves essential for recovering appearance cues that semantics alone cannot provide. We presented the ablation studies in Sec.~\ref{sec:ablation}. These datasets do not provide a unified explicit license but are commonly used for academic research and are properly cited. We follow their respective terms of use.

\subsection{Temporal adapter and distillation}

To equip the frozen diffusion backbone with temporal consistency, we described in Sec.~\ref{sec:temporal_adapter} a temporal adapter, which we implemented using a pretrained AnimateDiff model with video length $K=16$~\cite{guo2023animatediff}. We extend this model to causal autoregressive synthesis as detailed in Sec.~\ref{sec:autoregressive}. Here, we described a method to convert the bidirectional temporal attention into a causal, windowed variant and reduce the number of diffusion steps. DMD was shown to be effective in this context~\cite{yin2025slow}. We apply DMD, however, to distill only the Temporal Adapter parameters $\phi$ of the model $\epsilon_{\theta,\phi}$ while keeping the diffusion backbone fixed. Our approach allows the temporal component to be trained in isolation, substantially reducing training cost compared to full-model distillation. The student is initialized using Latent Consistency Models (LCM) with a window size of $W=3$, sampling $10^3$ videos from the teacher, followed by $8\times10^3$ LCM update steps~\cite{luo2023latent}. After this initialization, DMD is performed for $50\times10^3$ steps on the OpenVid-1M dataset~\cite{nan2024openvid}, using a batch size of 6 (effective $6\times4 = 24$). The student is trained to minimize the $\mathcal{L}_{DMD}$ in Eq.~\ref{eq:dmd}, and the critic is trained to minimize the diffusion loss $\mathcal{L}_D$ in Eq.~\ref{eq:eps_loss}. The student was trained to denoise from the fixed diffusion timesteps of $t \in \{ 999, 759, 499, 259 \}$. As shown in Fig.~\ref{fig:distillation}, the student employs causal while the critic uses bidirectional attention. OpenVid-1M is licensed under CC-BY 4.0 and is used in accordance with its terms for research purposes.

\subsection{Final model}

After training the three modules, namely the Semantic Control, Restoration Adapter, and Temporal Adapter, separately, we combine them into the model $\epsilon_{\theta,\omega,\psi,\phi^-}(x_t^{k-3:k}, \tilde{x}_t^{k-3:k}, Y^{k-3:k}, t)$ and extend it with KV-caching to enable efficient autoregressive synthesis. When run on an NVIDIA RTX 4090 at a resolution of $512\times512$ with $T=4$ diffusion steps and a window size of $W=3$, our approach achieves, without any special optimizations, $5.1$ FPS while using $6.2$ GB of VRAM. This is not real-time. But paves the way to real-time autoregressive and causal ultra-low-bitrate video reconstruction.

\section{Temporal-only distillation}

\begin{figure}[!t]
\centering
\includegraphics[width=0.4\linewidth]{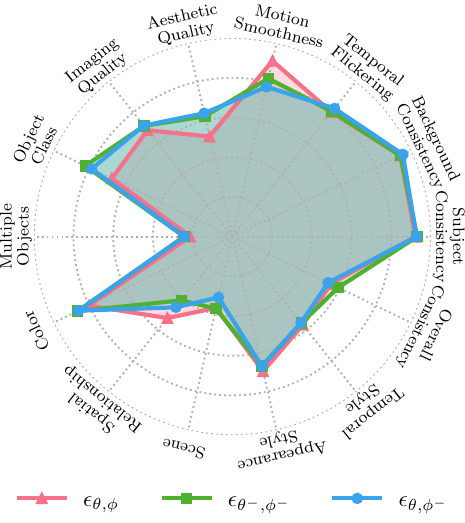}
\caption{Radar plot of the VBench text-to-video evaluation comparing the teacher model $\epsilon_{\theta,\phi}$, the distillation of backbone and Temporal Adapter $\epsilon_{\theta^-,\phi^-}$, and our Temporal Adapter-only distillation $\epsilon_{\theta,\phi^-}$. All perform comparably.}
\label{fig:radar}
\end{figure}

As discussed in Sec.~\ref{sec:autoregressive}, we distill only the Temporal Adapter parameters $\phi$, rather than the full model $\epsilon_{\theta,\phi}$, which substantially reduces the number of trainable parameters and enables an efficient distillation process. To evaluate this approach, we compare the temporal-only distilled model $\epsilon_{\theta,\phi^-}$ with both the non-distilled teacher $\epsilon_{\theta,\phi}$ and a fully distilled variant $\epsilon_{\theta^-,\phi^-}$. Since this analysis focuses purely on the temporal behavior, we treat it as a text-to-video synthesis task and employ the VBench evaluation~\cite{huang2023vbench}. This experiment highlights how isolating the temporal module enables efficient training while preserving the performance for causal and few-step synthesis. Fig.~\ref{fig:radar} compares the performance, and Tab.~\ref{tab:distill} reports the required training resources.

The radar plot in Fig.~\ref{fig:radar} shows that all three models achieve similar performance. However, as summarized in Table~\ref{tab:distill}, our temporal-only distillation requires substantially fewer training resources and provides almost twice the training speed of a fully distilled model, while achieving similar VBench scores. The teacher model performs similarly but cannot be used in a causal, real-time video reconstruction setting. This large-scale evaluation already establishes the effectiveness and efficiency of selectively distilling the temporal module.

\begin{table}
\centering
\caption{Comparison of full and temporal-only distillation. Our approach requires significantly fewer training resources. \textbf{$300\times$} reduction in trained parameters and \textbf{$2\times$} reduction in GPU hours.}
{\footnotesize
\begin{tabular}{cccccccc}
    \toprule
    & & & & & \multicolumn{3}{c}{VBench} \\
    \cmidrule(lr){6-8}
    \makecell{Method} &
    \makecell{Trained parameters} &
    \makecell{Batch} &
    \makecell{Accumulation} &
    \makecell{GPU Hours} &
    Quality &
    Semantic &
    Overall \\
    \midrule
    $\epsilon_{\theta^-,\phi^-}$ & $1.3 \times 10^9$ & $2$ & $3$ & $12.5 \times 4$ & 0.8196 & 0.6144 & 0.7786 \\
    $\epsilon_{\theta,\phi^-}$   & $4.1 \times 10^6$ & $6$ & $1$ & $6 \times 4$    & 0.8206 & 0.6019 & 0.7768 \\
    \bottomrule
\end{tabular}
}
\label{tab:distill}
\end{table}

\begin{figure}[!t]
 \centering
    \subfloat[Cityscapes]{%
        \includegraphics[width=0.48\linewidth]{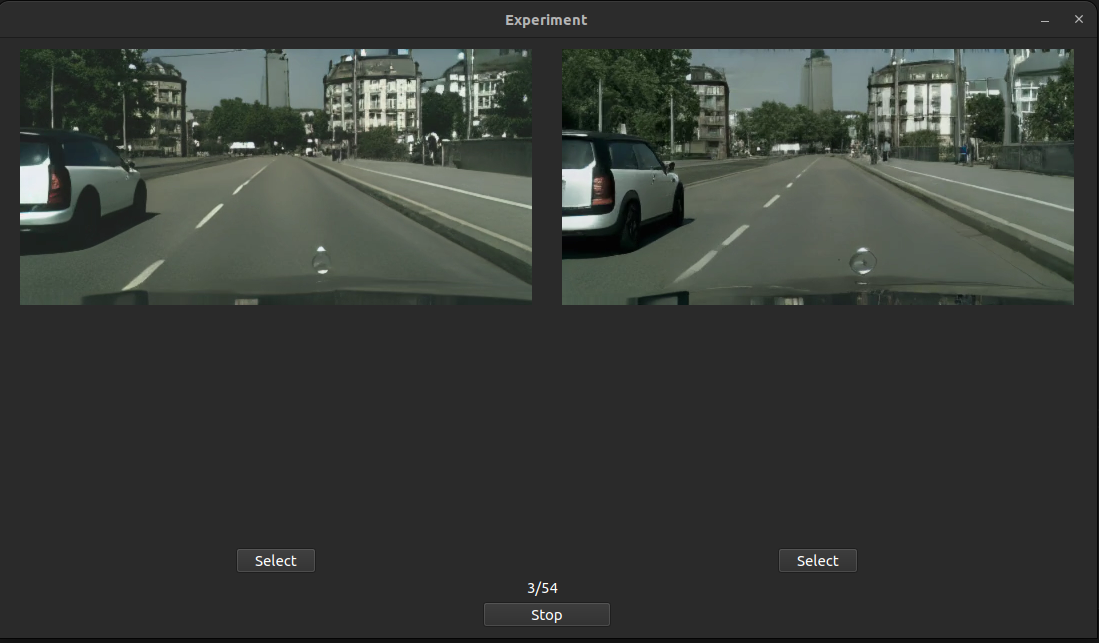}%
    }
    \hfill
    \subfloat[YCB-Sim]{%
        \includegraphics[width=0.48\linewidth]{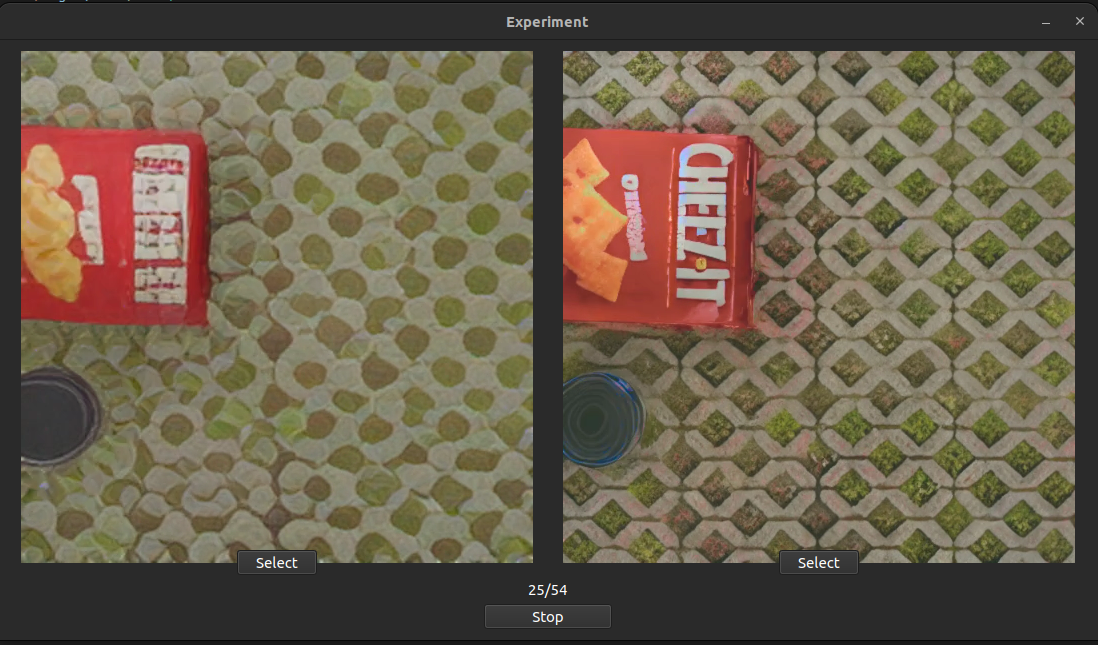}%
    }
\caption{Screenshots from subjective evaluation GUI.}
\label{fig:gui}
\end{figure}

\section{Subjective evaluation} \label{sec:subjective_details}

We conducted a randomized pairwise forced-choice user study to assess perceptual quality. In each trial, participants viewed two anonymized reconstructions of the same source video in a side-by-side interface. Our method was compared against a single baseline, and participants selected the result they preferred based on overall visual quality. The left/right placement of our method was randomized and counterbalanced across trials to mitigate positional bias. Fig.~\ref{fig:gui} displays a screenshot from the subjective experiment GUI. Participation was voluntary, and no compensation was provided.

We sampled $72$ source videos from each dataset (Cityscapes and YCB-Sim) and compared our method against all available baselines. This sampling was uniform and did not depend on any metric. Furthermore, these samples were not included in the hyperparameter analysis done in Sec.~\ref {sec:hyperparameters}. As discussed above, the baseline EVC was not evaluated in YCB-Sim because no training script was available. For Cityscapes, this yielded $72 \times 5 = 360$ pairwise comparisons; for YCB-Sim, $72 \times 4 = 288$. Each comparison was evaluated by two independent participants, resulting in a total of $2 \times (360 + 288) = 1,296$ assessments.

Comparison assignments were precomputed to balance participant workload, dataset coverage, and baseline frequency, while ensuring that no participant saw the same source video more than once to avoid bias. We recruited $24$ participants, corresponding to $54$ assessments per participant. The subjective study was performed under ethical committee approval and with informed consent with the following script

\begin{mdframed}[backgroundcolor=gray!15, linewidth=0pt]
    Thank you for your participation. On the screen in front of you, you will be presented with two reconstructions of a video. We ask that you compare the visual quality of these two looping videos and select the one that appeals to you most. Once the selection is done, new videos will appear. There is no specific definition of quality; we are interested in your subjective choice. There is no wrong answer. We do not measure the time nor collect any data other than your choice. You will evaluate 54 videos in total. The experiment will last approximately 15 minutes. The videos will be presented from two distinct sets. In one, you will observe dash cam videos of a car driving, and in the other, you will observe objects moving on a tabletop. You can adjust the monitor and the chair according to your needs. No risks are associated with participation in this study. If you feel uncomfortable at any point, you can stop the experiment. 
\end{mdframed}

For each baseline, we report preference rates with $95\%$ Wilson confidence intervals. Statistical significance was evaluated using one-sided binomial tests under the null hypothesis that both methods were equally likely to be preferred ($0.5$). To account for multiple baseline comparisons, p-values were adjusted using the Holm procedure.

\section{Baseline details} \label{sec:baseline_details}

As discussed in Sec.~\ref{sec:baselines}, we compare our method against several baselines. This section describes the implementation and experimental settings used for each method.

\subsection{VVC}

We use the open-source implementation of the VVC encoder\footnote{\url{https://github.com/fraunhoferhhi/vvenc}} and decoder\footnote{\url{https://github.com/fraunhoferhhi/vvdec}} with the medium preset. Bitrate is controlled by the quantization parameter $q$. For ultra-low-bitrate settings where the target bitrate cannot be reached at the original resolution, we downsample the input video; otherwise, the original resolution is used. The intra-period is set to $32$. In Tab.~\ref{tab:rate_params_vvc} we provide the VVC bitrate parameters.

\begin{table}[h!]
\centering
\caption{Bitrate control parameters of VVC.}
\scriptsize{
    \begin{tabular}{@{}ccccccccc@{}}
    \toprule
     & $q$ & $\downarrow$ & $q$ & $\downarrow$ & $q$ & $\downarrow$ & $q$ & $\downarrow$ \\ \midrule
    Cityscapes & 40 & \multicolumn{1}{c|}{2} & 45 & \multicolumn{1}{c|}{2} & 50 & \multicolumn{1}{c|}{2} & 55 & 2 \\
    bpp & \multicolumn{2}{c|}{0.0143} & \multicolumn{2}{c|}{0.0085} & \multicolumn{2}{c|}{0.005} & \multicolumn{2}{c}{0.0035} \\ \midrule
    YCB & 45 & \multicolumn{1}{c|}{4} & 50 & \multicolumn{1}{c|}{4} & 55 & \multicolumn{1}{c|}{4} & 60 & 4 \\
    bpp & \multicolumn{2}{c|}{0.0023} & \multicolumn{2}{c|}{0.0015} & \multicolumn{2}{c|}{0.0012} & \multicolumn{2}{c}{0.001} \\ \bottomrule
    \end{tabular}
}
\label{tab:rate_params_vvc}
\end{table}

\subsection{DCVC}

We use the open-source DCVC-RT implementation\footnote{\url{https://github.com/microsoft/dcvc}}, a state-of-the-art neural video codec. The bitrate is controlled via the quality parameter $q$. As in VVC, we downsample the input video if the target bitrate cannot be achieved at the original resolution. The intra-period is set to $32$. DCVC is additionally used to generate degraded ultra-low-bitrate inputs for our method. Further coding details are provided in Sec.~\ref{sec:rate_settings}. In Tab.~\ref{tab:rate_params_dcvc} we provide the DCVC bitrate parameters.

\begin{table}[h!]
\centering
\caption{Bitrate control parameters of DCVC.}
\scriptsize{
    \begin{tabular}{@{}ccccccccc@{}}
    \toprule
     & $q$ & $\downarrow$ & $q$ & $\downarrow$ & $q$ & $\downarrow$ & $q$ & $\downarrow$ \\ \midrule
    Cityscapes & 0 & \multicolumn{1}{c|}{1} & 5 & \multicolumn{1}{c|}{1} & 10 & \multicolumn{1}{c|}{1} & 15 & 1 \\
    bpp & \multicolumn{2}{c|}{0.0045} & \multicolumn{2}{c|}{0.006} & \multicolumn{2}{c|}{0.0081} & \multicolumn{2}{c}{0.011} \\ \midrule
    YCB & 5 & \multicolumn{1}{c|}{2} & 10 & \multicolumn{1}{c|}{2} & 15 & \multicolumn{1}{c|}{2} & 20 & 2 \\
    bpp & \multicolumn{2}{c|}{0.0009} & \multicolumn{2}{c|}{0.0013} & \multicolumn{2}{c|}{0.0018} & \multicolumn{2}{c}{0.0024} \\ \bottomrule
    \end{tabular}
}
\label{tab:rate_params_dcvc}
\end{table}

\subsection{EVC}

We use the inference code provided by the authors of EVC\footnote{\url{https://github.com/ming-liuyi/Extreme-Video-Compression-With-Prediction-Using-Pre-trainded-Diffusion-Models-}} along with their pretrained checkpoint trained on Cityscapes. Since the training code is not available, we do not report results on YCB-Sim. Bitrates are controlled via pretrained model quality levels $q=\{2,3,4,5\}$ and an LPIPS threshold of $\tau=0.25$ for intra frames. In Tab.~\ref{tab:rate_params_evc} we provide the EVC bitrate parameters.

\begin{table}[h!]
\centering
\caption{Bitrate control parameters of EVC.}
\scriptsize{
    \begin{tabular}{@{}ccccccccc@{}}
    \toprule
     & $q$ & $\tau$ & $q$ & $\tau$ & $q$ & $\tau$ & $q$ & $\tau$ \\ \midrule
    Cityscapes & 2 & \multicolumn{1}{c|}{0.25} & 3 & \multicolumn{1}{c|}{0.25} & 4 & \multicolumn{1}{c|}{0.25} & 5 & 0.25 \\
    bpp & \multicolumn{2}{c|}{0.0047} & \multicolumn{2}{c|}{0.006} & \multicolumn{2}{c|}{0.0081} & \multicolumn{2}{c}{0.012} \\ \bottomrule
    \end{tabular}
}
\label{tab:rate_params_evc}
\end{table}

\subsection{GLC}

We use the open-source implementation of GLC Video\footnote{\url{https://github.com/jzyustc/GLC}}, extending the code to report the real bitrate instead of the theoretical as in \texttt{image\_model.py\#L136} (function \texttt{get\_y\_gaussian\_bits}). Videos are encoded using arithmetic coding with DCVC-style headers to ensure consistency with other methods. Similar to VVC and DCVC, we downsampled the videos if the target bitrates were not reachable. In Tab.~\ref{tab:rate_params_glc} we provide the GLC bitrate parameters.

\begin{table}[h!]
\centering
\caption{Bitrate control parameters of GLC.}
\scriptsize{
   \begin{tabular}{@{}ccccccccc@{}}
    \toprule
     & $q$ & $\downarrow$ & $q$ & $\downarrow$ & $q$ & $\downarrow$ & $q$ & $\downarrow$ \\ \midrule
    Cityscapes & 50 & \multicolumn{1}{c|}{2} & 20 & \multicolumn{1}{c|}{1.5} & 0 & \multicolumn{1}{c|}{1.25} & 10 & 1 \\
    bpp & \multicolumn{2}{c|}{0.0066} & \multicolumn{2}{c|}{0.0094} & \multicolumn{2}{c|}{0.0102} & \multicolumn{2}{c}{0.0124} \\ \midrule
    YCB & 40 & \multicolumn{1}{c|}{8} & 20 & \multicolumn{1}{c|}{4} & 10 & \multicolumn{1}{c|}{3} & 20 & 2 \\
    bpp & \multicolumn{2}{c|}{0.0012} & \multicolumn{2}{c|}{0.0015} & \multicolumn{2}{c|}{0.0019} & \multicolumn{2}{c}{0.0026} \\ \bottomrule
    \end{tabular}
}
\label{tab:rate_params_glc}
\end{table}

\subsection{I2V-SC}

We reproduce the I2V-SC baseline using the Image-to-Video Adapter~\cite{i2v} and its open-source implementation\footnote{\url{https://github.com/KlingAIResearch/I2V-Adapter}}. This baseline shares the same diffusion backbone, allowing a direct comparison and extension with our ControlNet-based Semantic Control and AnimateDiff-based causal Temporal Adapter described in Sec.~\ref{sec:implementation}. This method operates solely on the ultra-low-bitrate semantic representations described in Sec.~\ref{sec:semantic_video_coding}. In this context, the parameters $P$ and $\xi$ control the bitrate. In Tab.~\ref{tab:rate_params_i2v} we provide the semantic-only bitrate parameters. Notice that this model cannot achieve higher rates in YCB-Sim, as the contour complexity is low.

\begin{table}[h!]
\centering
\caption{Bitrate control parameters of I2V-SC.}
\scriptsize{
    \begin{tabular}{@{}ccccccccc@{}}
\toprule
 & $P$ & $\xi$ & $P$ & $\xi$ & $P$ & $\xi$ & $P$ & $\xi$ \\ \midrule
Cityscapes & 8 & \multicolumn{1}{c|}{4} & 4 & \multicolumn{1}{c|}{6} & 2 & \multicolumn{1}{c|}{6} & 2 & 4 \\
bpp & \multicolumn{2}{c|}{0.0053} & \multicolumn{2}{c|}{0.0062} & \multicolumn{2}{c|}{0.0089} & \multicolumn{2}{c}{0.0113} \\ \midrule
YCB & 1 & \multicolumn{1}{c|}{10} & 1 & \multicolumn{1}{c|}{8} & 1 & \multicolumn{1}{c|}{6} & 1 & 4 \\
bpp & \multicolumn{2}{c|}{0.001} & \multicolumn{2}{c|}{0.0011} & \multicolumn{2}{c|}{0.0012} & \multicolumn{2}{c}{0.0013} \\ \bottomrule
\end{tabular}
    }
\label{tab:rate_params_i2v}
\end{table}

\section{Broader impacts} \label{sec:impacts}

This work studies the reconstruction of videos from ultra-low-bitrate representations using generative diffusion models. Potential benefits include enabling efficient video transmission in bandwidth-constrained settings, such as remote sensing, telemedicine, mobile communication, and interstellar communication, where reduced bitrates can improve accessibility and scalability.

However, improvements in generative video reconstruction may also introduce risks. As with other generative models, the ability to synthesize realistic video content from limited input may be misused to generate misleading or manipulated visual content, potentially contributing to disinformation or fraud. In addition, reconstruction from sparse inputs may hallucinate plausible but incorrect details, which could be problematic in safety-critical applications. Hence, the proposed method is not intended for safety-critical applications such as driving.

% \newpage
% \input{checklist.tex}

\end{document}